\documentclass[runningheads,a4paper]{llncs}

\usepackage{amssymb}
\usepackage{graphicx}

\usepackage{color}
\usepackage{xcolor}
\usepackage{graphicx}
\usepackage{subfigure}
\usepackage{cite}
\usepackage{amsmath}
\usepackage{amsfonts}
\usepackage{enumerate}

\usepackage{algorithm}
\usepackage{algpseudocode}
\usepackage{tabulary}
\usepackage{multirow}

\usepackage{graphicx}
\usepackage{pgf}
\usepackage{pgfplots}
\usepackage{tikz}
\usepackage{color}
\usepackage{xxcolor}
\usetikzlibrary{arrows,shadows,petri,automata,shapes.geometric}
\usepgfplotslibrary{groupplots,fillbetween}
\usepackage{epstopdf}
\usepackage{filecontents}
\usepackage[misc]{ifsym}

\usetikzlibrary{external}
\newtheorem{defn}{Definition}

\usepackage{url}
\urldef{\mailsa}\path|{marco.minelli, lorenzo.sabattini}@unimore.it|

\newcommand{\keywords}[1]{\par\addvspace\baselineskip
\noindent\keywordname\enspace\ignorespaces#1}

\begin{document}

\mainmatter  

\title{Stop, Think, and Roll: Online Gain Optimization for Resilient Multi-robot Topologies}

\titlerunning{Optimized control law for resilient multi-robot topologies}

%
%
\author{Marco Minelli$^1$\and Marcel Kaufmann$^2$ \and Jacopo Panerati$^2$\and Cinara Ghedini$^3$\and Giovanni Beltrame$^2$ \and Lorenzo Sabattini$^1$}
\authorrunning{Optimized control law for resilient multi-robot topologies}

\institute{$^1$Department of Sciences and Methods for Engineering, \\Universit\`a degli Studi di Modena e Reggio Emilia\\
\vspace{0.5em}
$^2$Department of Software and Computer Engineering,\\
Polytechnique Montr\'eal \\
\vspace{0.5em}
$^3$Departamento de Computa\c{c}\~ao Cient\'ifica,\\
Instituto Tecnol\'ogico de Aeron\'autica
}

%
%

\toctitle{ }
\tocauthor{ }
\maketitle

\begin{abstract}
Efficient networking of many-robot systems is considered one of the grand challenges of robotics. In this article, we address the problem of achieving resilient, dynamic interconnection topologies in multi-robot systems. In scenarios in which the overall network topology is constantly changing, we aim at avoiding the onset of single points of failure, particularly situations in which the failure of a single robot causes the loss of connectivity for the overall network. We propose a method based on the combination of multiple control objectives and we introduce an online distributed optimization strategy that computes the optimal choice of control parameters for each robot. This ensures that the connectivity of the multi-robot system is not only preserved but also made more resilient to failures, as the network topology evolves. We provide simulation results, as well as experiments with real robots to validate theoretical findings and demonstrate the portability to robotic hardware.
%
%
%
%

\keywords{resilience; multi-robot systems; connectivity; graph theory; control; online optimization; robotic hardware; experimental validation.}
\end{abstract}

\section{Introduction}\label{sec:intro}

Robot swarms are defined as one of the grand challenges of Science Robotics~\cite{yang2018grand}, and resilient networking in particular is a key technology for their successful implementation and deployment. Yang et al.~\cite{yang2018grand} define resilience as a ``property that is about systems that can bend without breaking. Resilient systems are self-aware and self-regulating, and can recover from large-scale disruptions''. In this paper, we study the problem of optimizing the resilience of the connectivity of a dynamic multi-robot network, with respect to robotic failures. 

%
When looking at multi-robot and swarm systems, we can observe that they are inherently redundant. Being composed of multiple entities, failure of a single robot does not, in general, cause the complete failure of the multi-robot system. It might still be able to complete its objectives, possibly with a decreased performance by exhibiting graceful degradation.

To cooperate and achieve shared objectives, robots need to exchange information among each other. When considering groups of mobile robots with limited-range communication capabilities, the topology of the robot network changes as the robots move. It is then imperative to impose constraints on the robots' motion, so that connectivity is preserved. Several strategies can be found in the literature to address the connectivity preservation problem~\cite{Yang:2010,Sabattini:2013,sabattini2013b,gasparritro2017,poonawala2015,ji2007}. 

%
It is worth noting that, when considering real robotic systems, connectivity preservation is not always a sufficient safeguard for network resilience. In general, robots can be prone to failures and in many interconnection topologies, individual failures can generate the immediate loss of connectivity of the overall network, possibly preventing the multi-robot system from achieving the desired cooperative results. The inherent reliability of multi-robot systems (due to redundancy) is then heavily reduced by the presence of single points of failure.

To address this issue, Ghedini et al. propose a control strategy~\cite{ghedini2017b} to ensure that a multi-robot system preserves a high level of connectivity, while performing its assigned task. The proposed methodology is based on the combination of multiple control laws aiming at adjusting the interconnection topology when potentially vulnerable topological configurations 
are identified. In~\cite{panerati2018}, this method was implemented on a real multi-robot system, controlled to perform area coverage, and its performance was evaluated against the shortcomings of the real-world, such as imperfect communication. Again, this is achieved through the combination of different control laws, each associated to a gain. These gain combinations determine the overall performance of the system. To automate the choice of the overall parameter set (i.e., the control law gains), we introduce an offline optimization method~\cite{panerati2018}.

The main drawback of this approach is that the chosen gains are optimal values obtained by averaging metrics over a set of different topologies, therefore losing the advantage of optimizing for the current topology of the network. This is particularly significant for a mobile robot network, where the topology changes dynamically as the robots move.

Consequently, in this paper, we propose an online optimization strategy, that allows the multi-robot system to estimate, at run time, the optimal (or close to optimal) set of control law gains that optimizes the overall performance of the system.

The rest of this paper is organized as follows. The necessary background on network properties is presented in Section~\ref{evaluationMechanisms}. The system model and the problem addressed here are discussed in Section~\ref{sec:problem}. Section~\ref{sec:overview} outlines the control architecture. The optimized control strategy is described in details in Section \ref{sec:optimization}, and the experimental validation on real set-up in presented in Section ~\ref{sec:experiments}. Finally, Section~\ref{sec:conclusions} discusses the future directions and concludes the document.

\section{Preliminaries: network properties} \label{evaluationMechanisms}
%
%


%
Consider an undirected graph $\mathcal{G}$, where $\mathcal{V}\left(\mathcal{G}\right)$ and $\mathcal{E}\left(\mathcal{G}\right) \subset \mathcal{V}\left(\mathcal{G}\right) \times \mathcal{V}\left(\mathcal{G}\right)$ are the vertex set and the edge set, respectively.
Moreover, let $W \in \mathbb{R}^{N \times N}$ be the weight matrix: each element $w_{ij}$ is a positive number if an edge exists between nodes $i$ and $j$, zero otherwise. Since $\mathcal{G}$ is undirected, then $w_{ij}=w_{ji}$.

Thus, let $\mathcal{L}\in\mathbb{R}^{N\times N}$ be the Laplacian matrix of graph $\mathcal{G}$ and $D=\mbox{diag}\left(\left\{k_i\right\}\right)$ be the degree matrix, where $k_i$ is the degree of the $i$-th node of the graph, i.e. \mbox{$k_i=\sum\limits_{j=1}^{N}{w_{ij}}$}. The (weighted) Laplacian matrix of the graph is then defined as \mbox{$L=D-W$}. 
As it is well known from algebraic graph theory, the Laplacian matrix of an undirected graph $\mathcal{G}$ exhibits some remarkable properties regarding its connectivity \cite{algebraicgraphtheory2001}. Let $\lambda_i$, $i=1,\ldots,N$ be the eigenvalues of the Laplacian matrix, then:	
\begin{itemize}
	\item The eigenvalues are real, and can be ordered such that $0=\lambda_1\leq\lambda_2\leq\ldots\leq\lambda_N$
	\item Define now $\lambda = \lambda_2$. Then, $\lambda>0$ if and only if the graph is connected. Therefore,
	$\lambda$ is defined as the \textbf{algebraic connectivity} of the graph: in 
	a weighted graph, $\lambda$ is a non-decreasing function of each edge weight.
\end{itemize}

Connected graphs may get disconnected in case of failure of one or more nodes. Different nodes have different roles in maintaining the overall network connectivity. 
The concept of \emph{centrality} is usually exploited to identify the most important nodes within a graph \cite{koschutzki2005}. 
In particular, referring to connectivity maintenance, we will consider the concept of \emph{Betweenness Centrality} ($BC$) \cite{citeulike:165614}, which establishes higher scores for nodes that are contained in most of the shortest paths between every pair of nodes in the network. 
%
%
%
According to this definition of centrality, removing the most central nodes might quickly lead to network fragmentation. We, therefore, introduce the following definition of the \emph{Robustness level}.
\begin{defn}[Robustness level~\cite{syroco2015c}]\label{robustness_level}
	Consider a graph $\mathcal{G}$ with $N$ nodes. Let  $\left[v_1, \ldots, v_N\right]$ be the list of nodes ordered by descending value of $BC$. Let $\varphi <N$ be the minimum index $i \in \left[ 1, \ldots, N\right]$ such that, removing nodes $\left[v_1, \ldots, v_i\right]$ leads to disconnecting the graph, that is, the graph including only nodes $\left[v_{\varphi+1}, \ldots, v_N\right]$ is disconnected.
	Then, the network \emph{robustness level} of $\mathcal{G}$ is defined as:
	\begin{equation} \label{eq:robustness}
		\Theta(\mathcal{G}) =  \frac{\varphi}{N}
	\end{equation}
\end{defn}

The robustness level defines the fraction of central nodes that need to be removed from the network to obtain a disconnected network. 
Small values of $\Theta(\mathcal{G})$ imply that a small fraction of node failures may fragment the network. Therefore, increasing this value means increasing the network resilience, that is, its robustness to failures.
We observe that $\Theta(\mathcal{G})$ is only an estimate of how far the network is from getting disconnected w.r.t. fraction of nodes removed. In fact, it might be the case that different orderings of nodes with the same $BC$ produce different values of $\Theta(\mathcal{G})$.

While the robustness level refers to the overall state of the network, from a local perspective, a heuristic for estimating the vulnerability level of a node by means of the information acquired from its 1-hop and 2-hops neighbors was proposed in \cite{syroco2015c}. We summarize it as follows: let $d(v,u)$ be the shortest path between nodes $v$ and $u$, i.e., the minimum number of edges that connect nodes $v$ and $u$. Subsequently, define $\Pi(v)$ as the set of nodes from which $v$ can acquire information:
\begin{equation}
	\Pi(v)=\{u \in V(G) : d(v,u)\leq 2\}
\end{equation}
Moreover, let $|\Pi(v)|$ be the number of elements of $\Pi(v)$. In addition, define $\Pi_{2}(v)\subseteq \Pi(v)$ as the set of the 2-hop neighbors of $v$, that comprises only nodes whose shortest path from $v$ is exactly equal to 2 hops, namely:
\begin{equation}
	\Pi_{2}(v)=\{u \in V(G) : d(v,u)=2\}
\end{equation}

Now define $L(v,u)$ as the \emph{number of paths} between nodes $v$ and $u$, and let $Path_{\beta}(v) \subseteq \Pi_{2}(v)$ be the set of $v$'s 2-hop neighbors that are reachable through at most $\beta$ paths, namely:
\begin{equation}
	Path_{\beta}(v)=\{u \in \Pi_{2}(v): L(v,u) \leq \beta\}
\end{equation}
Thus, $\beta$ defines the threshold for the maximal number of paths between a node $v$ and each of its $u$ neighbors that are necessary to include $u$ in $Path_{\beta}(v)$. Therefore, using a low value for $\beta$, it is possible to identify the most weakly connected 2-hop neighbors. Hence, the value of $|Path_{\beta}(v)|$  is an indicator of the magnitude of node fragility w.r.t. connectivity, and \textbf{the vulnerability level of a node regarding failures} is given by $P_{\theta}(v) \in \left(0,1\right)$:

\begin{equation} \label{eq:vulnerability}
	P_{\theta}(v) =  \frac{|Path_{\beta}(v)|}{|\Pi(v)|}
\end{equation}

We will hereafter use $\beta=1$, in order to identify 2-hop neighbors that are connected by a single path, which can represent a critical situation for network connectivity.

\section{System model and problem formulation}\label{sec:problem}

We consider a multi-robot system composed of $N$ mobile robots and we assume that each robot is able to communicate with other robots within a communication radius $R$. 
The resulting communication topology is represented by an undirected graph $\mathcal{G}$.

Let the state of each robot be its position $p_i\in\mathbb{R}^m$, and
let $p=\left[p_1^T\ldots p_N^T\right]^T\in\mathbb{R}^{N m}$ be the
state vector of the multi-robot system. Let each robot be modeled as a single
integrator system, whose velocity can be directly controlled:
\begin{equation}
	\dot{p}_i = u_i
	\label{eq:singleintegrator}
\end{equation}
where $u_i\in \mathbb{R}^m$ is a control input.
%
%
For each robot, the control input has to be defined so that a global objective can be achieved. As an example of a commonly implemented application, in the rest of the paper, we will refer to a scenario in which the robots are controlled to spread in a given area while avoiding collisions. However, the proposed methodology can be easily extended to other coordinated control objectives.


It is worth noting that coordinated objectives can be achieved only if information can be exchanged among the robots, that is if the communication graph is connected, and the robots keep this property as the system evolves. However, when considering real robotic systems, failures can not be neglected: robots can stop working unexpectedly and become unable to collaborate. 


In this paper we combine different control laws, guaranteeing the achievement of a common objective (area coverage, in our case) while ensuring the preservation of the connectivity for the communication graph, even in the presence of failures. The combination of the different control laws aims at maximizing a global performance index. This index defines a trade-off between the area actually covered by the robots, and the level of connectivity of the communication network.

\section{Overview of the control architecture}\label{sec:overview}


Referring to the kinematic model in Equation~\eqref{eq:singleintegrator}, in the following, we consider each robot to be controlled by means of a control input defined as the superposition of three different terms, that is:
\begin{equation}
	u_i=\sigma u_i^c + \psi u_i^r + \zeta u_i^{d}
	\label{eq:combinedinput}
\end{equation}
The contributions that constitute the control inputs are defined as follows:
\begin{itemize}
	\item The term $u_i^c\in\mathbb{R}^m$ represents the connectivity preservation control input. The role of this control input is to enforce that, if the communication graph is initially connected, then it will remain connected as the system evolves.
	\item The term $u_i^r\in\mathbb{R}^m$ represents the topology resilience improvement control input. This term aims at improving the robustness of the topology against failures. In other words, its purpose is to minimize the presence of single points of failure that could induce a disconnection in the communication graph in case of failure of one or more robots.
	\item The term $u_i^d\in\mathbb{R}^m$ represents the desired control action. This encodes the coordinated objective that the multi-robot system needs to achieve. As a representative example, in this paper, we consider the objective to be the uniform coverage of a given area.
	\item The terms $\sigma, \psi, \zeta \geq 0$ represent linear combination gains. They define the relative importance of the separate control laws.
\end{itemize}

It is worth noting that the overall behavior of the multi-robot system is defined by the way in which each individual control action is defined, and by how they are combined. Indeed, a different choice of the linear combination gains leads to a different behavior of the multi-robot system.

In the following subsections we will introduce representative examples of the individual control actions, that we will consider for implementation in the rest of the paper. 

\subsection{Connectivity preservation}  \label{sec:connectivityModel}

We consider the connectivity preservation control term $u_i^c$ to be designed, as in~\cite{sabattiniijrr2013}, to ensure that the value of the algebraic connectivity $\lambda$ never goes below a given threshold $\epsilon>0$. As in~\cite{sabattiniijrr2013}, the following \emph{energy function} can be used for generating the decentralized connectivity maintenance control strategy:
\begin{equation}
	V\left(\lambda\right)=\left\{
	\begin{array}{ll}
		\coth \left(\lambda - \epsilon\right)\,\, & \mbox{if } \lambda>\epsilon\\
		0 & \mbox{otherwise.}
	\end{array}\right.
	\label{eq:totaltensionCDC}
\end{equation}
The control law is then designed to drive the robots to perform a gradient descent of $V\left(\cdot\right)$, which ensures preservation of the graph connectivity. 
Considering the robot model introduced in~\eqref{eq:singleintegrator}, the control law is defined as follows:
\begin{equation}
	{u}_i=u_i^c = -\dfrac{\partial V\left(\lambda\right)}{\partial p_i}=-\dfrac{\partial V\left(\lambda\right)}{\partial \lambda}\dfrac{\partial \lambda}{\partial p_i}.
	\label{eq:Kfunction}
\end{equation}
We observe that the connectivity preservation framework can be enhanced to consider also additional objectives. In particular, as shown in~\cite{robuffogiordano2013}, the concept of \emph{generalized connectivity} can be utilized to simultaneously guarantee \textbf{connectivity maintenance and collision avoidance} with environmental obstacles and among the robots.

\subsection{Topology resilience improvement} \label{sec:combinedModel}

We consider the topology resilience improvement control term $u_i^r$ to be designed---in accordance with the methodology defined in~\cite{adhoc2018,ghedini2017b}---to drive the robots toward an improved resilience of the interconnection topology.

Based on the concept of vulnerability level introduced in~\eqref{eq:vulnerability}, this  control strategy aims at increasing the number of links of a potentially vulnerable node $i$ towards its 2-hop neighbors that are in $Path_{\beta}(i)$, for a given value of $\beta$. Let $x_\beta^i\in\mathbb{R}^m$ be the barycenter of the positions of the robots in $Path_{\beta}(i)$. 
%
%
%
Considering the robot model introduced in~\eqref{eq:singleintegrator}, the control law is defined as follows:
\begin{equation}
	u_i^r = \xi_i\dfrac{x_\beta^i - p_i}{\left\| x_\beta^i - p_i \right\|}\alpha\left(t\right),
	\label{eq:controllaw}
\end{equation}
where $\alpha\left(t\right)\in\mathbb{R}$ is the linear velocity of the robots\footnote{Pathological situations may exist in which~\eqref{eq:controllaw} is not well defined, namely when $p_i = x_\beta^i$. However, this corresponds to the case where the $i$-th robot is exactly in the barycenter of its weakly connected 2-hop neighbors: in practice, this never happens when a robot detects itself as vulnerable. }. 

Parameter $\xi_i$ takes into account the vulnerability state of a node $i$, i.e., $\xi_i=1$ if node $i$ identifies itself as vulnerable or $\xi_i=0$ otherwise. As in~\cite{adhoc2018,ghedini2017b}, we set as vulnerable those robots $i$ exhibiting high values for $P_{\theta}(i)$: then, $\xi_i$ is defined as follows
\begin{equation}
	\xi_i=\left\{
	\begin{array}{ll}
		1 \quad & \text{if } P_{\theta}(i)>r\\
		0 \quad & \text{otherwise,}
	\end{array}\right.
	\label{eq:probabilisticparam}
\end{equation}
where $r\in(0,1)$ is a random number drawn from a uniform distribution, 
i.e., if $P_{\theta}(i)>r$, then the $i$-th robot considers itself as vulnerable. It is worth remarking that, according to~\eqref{eq:vulnerability}, each robot can evaluate its vulnerability level in a decentralized manner.

To summarize, this control law drives the vulnerable robots towards the barycenter of the robots in their $Path_{\beta}$, thus decreasing their distance to them, thus eventually creating new edges in the communication graph.

\subsection{Area coverage and collision avoidance}
To control the robot to evenly spread over a given area while avoiding collisions, we propose to use the well-known control strategy based on the Lennard-Jones potential~\cite{brambilla2013}.

At distance $x$ from its origin, the potential and the desired control action equations are:
\begin{equation}
	P_{LJ} = \iota \left(  \left( \frac{\delta}{x} \right)^{a} - 2 \cdot  \left( \frac{\delta}{x} \right)^{b}  \right)   ;\qquad
	u_i^d = -\iota \left(  \left( \frac{a\cdot \delta^a}{x^{a+1}} \right)^{a} - 2 \cdot  \left( \frac{b \cdot \delta}{x^{b+1}} \right)^{b}  \right)  
	\label{eq:lj-f}
\end{equation}
\noindent Parameters $\iota$ and $\delta$ represent the depth and distance
from the origin of the potential's minimum, respectively.  Exponents $a$ and
$b$ are set to 4 and 2.  For the sake of collision avoidance, we set $\delta$ to be larger than the
communication range of the robots. %

\section{Optimized control strategy}\label{sec:optimization}

In this section, we introduce a new methodology to achieve online optimization of the linear combination gains $\sigma$, $\psi$, $\zeta$ introduced in~\eqref{eq:combinedinput}. The objective is to make the robots able to identify the \emph{best} set of parameters during the evolution of the system.

The \emph{best} solution is defined starting from the system-level objective we are considering, that is, achieving area coverage while keeping a sufficiently high level of connectivity. For this purpose, we define the following objective function:
\begin{equation}
	f_{obj}(t) = \lambda(t)\mathcal{A}(t)
	\label{eq:objfun}
\end{equation}
where $\lambda(t)$ is the algebraic connectivity of the communication graph at time $t$, and $\mathcal{A}(t)$ is the value of the covered area at time $t$.

The choice of the objective function is motivated by the fact that this is 
a fundamentally multi-objective problem of two inversely proportional 
functions at different scales. A common way of avoiding an adaptive 
normalization scheme for the two functions is to consider their 
product~\cite{Panerati2014}.
This approach favors solutions that lead to a trade-off between maximizing the algebraic connectivity and maximizing the covered area.

\subsection{Proposed optimization method} \label{subsec:optimizationMethod}

The proposed optimization method aims at finding the optimal combination of the gains $\sigma, \psi, \zeta$ such that the objective function introduced in~\eqref{eq:objfun} is maximized at the end of each iteration. 
It is worth noting that the objective function is a product of nonlinear contributes, being computed using the algebraic connectivity (which, being an eigenvalue of the Laplacian matrix, is inherently nonlinear) and the covered area (which is calculated considering circular overlapping areas covered by each robot and therefore, it is also nonlinear).

Consequently, we consider the use of optimization methods that are well suited for nonlinear problems. In particular, we adopted the following methods~\cite{avriel2003}:
\begin{enumerate}
	\item 
	The \textbf{default search} optimization provides a uniform and default screening of the variable domain space. The main advantage of this method is the accuracy of the solution, which can be freely defined if the system performance is independent from the computation time. 
	\item \textbf{Random search}. This stochastic method does not require the gradient of the objective function and can consider non continuous or non differentiable objective functions. The optimal set of parameters is found  randomly probing the domain space and selecting the set which returns the highest objective function value. Random search algorithms are typically used to achieve high computational speed at the cost of losing formal guarantees of optimality.
	\item The \textbf{augmented Lagrangian} optimization algorithm method. Suited for constrained optimization problems, it requires to first penalize the objective function, translate the constrained optimization problem to a series of unconstrained problems, and then adds a term designed to mimic a Lagrange multiplier and improve precision and convergence speed. This algorithm exploits the gradient of the objective function to get the optimal set of parameters. Since the gradient is hard to compute for nonlinear functions such as~\eqref{eq:objfun}, numerical differentiation is exploited.
	
\end{enumerate}  

\subsection{Implementation and evaluation}
\label{sec:impl_eval}

We compared the optimization methods in terms of quality of the solution and required computation resources. To achieve this, we implemented the following procedure: 
\begin{enumerate}
	\item At each time step, the positions of all the robots are shared with all the other robots, according to the protocol described in~\cite{panerati2018}. To achieve this in a reasonable time without a fully connected network we use a consensus implementation---\emph{virtual stigmergy}~\cite{pinciroli2016a}---that was shown to scale with thousands of robots.
	\item Based on the shared positions, each robot computes the output of its own control law~\eqref{eq:combinedinput}, and its local estimate of each other robot's control law.
	\item Every $O_p$ time steps, each robot runs the optimization process to find an updated set of gains to be used in~\eqref{eq:combinedinput} according to the following sub-steps:
	\begin{enumerate}
		\item Each robot transforms the positions and the control laws in its local reference frame.
		\item A set of $G_p$ different values of the gains is generated (according to the considered optimization method described in Section~\ref{subsec:optimizationMethod}).
		\item For each value of the gains, the robots compute the predicted position of all the robots at the subsequent time step, and evaluate the objective function introduced in~\eqref{eq:objfun}.
		\item The gains that provide the highest value of the objective function are selected.
	\end{enumerate}
\end{enumerate}
For evaluating the different optimization methods, we implemented the control law using the Buzz scripting language~\cite{pinciroli2016}, and run a set of simulations using the multi-physics environment of ARGoS~\cite{pinciroli2012}. The optimization framework was written as a C++ module which communicates, by means of a socket, with the simulation set-up.

We evaluate the performance of each optimization method considering a network of 8 robots and compare against a system constant gains, considering all the possible combinations of the following sets:
\begin{equation}
	\psi = \left\{0,1,2\right\}\quad \sigma = \left\{0,1,2\right\}\quad \zeta = \left\{0,1\right\}
	\label{eq:gainsets}
\end{equation}
The configurations assessed start from the same (randomly selected) initial topology and are compared using the aforementioned three optimization algorithms.

\begin{figure}[!htb]
	\includegraphics[width=\columnwidth]{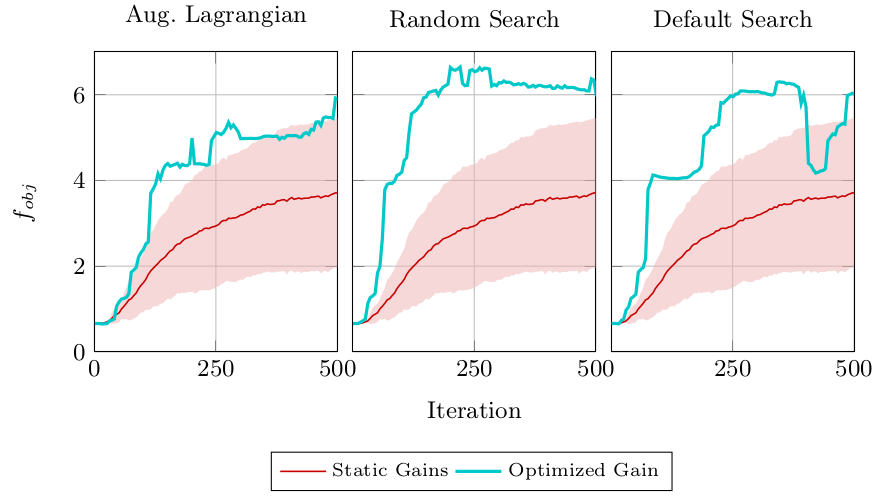}
	\caption{Objective function evolution comparison. Static gains simulations vs. optimized gains simulations for augmented Lagrangian, random, and default searches.}
	\label{opt1}       
\end{figure}

The results of the simulations are summarized in Fig.~\ref{opt1}, which depicts the value of the objective function~\eqref{eq:objfun} achieved with each optimization algorithm. In particular, the green line represents the objective function obtained with the optimization algorithm, while the red line with the corresponding shadow represents the average value and standard deviation of the objective function obtained with constant gains.

While in general the value of the objective function is typically greater when using the optimization method (with respect to constant gains), results show that \emph{random search} performs significantly better than other methods.
Furthermore, the computational requirements are generally smaller for \emph{random search}, in particular if compared with \emph{default search}, whose convergence time is approximately ten times larger.

\begin{figure}[!htb]
	\includegraphics[width=\columnwidth]{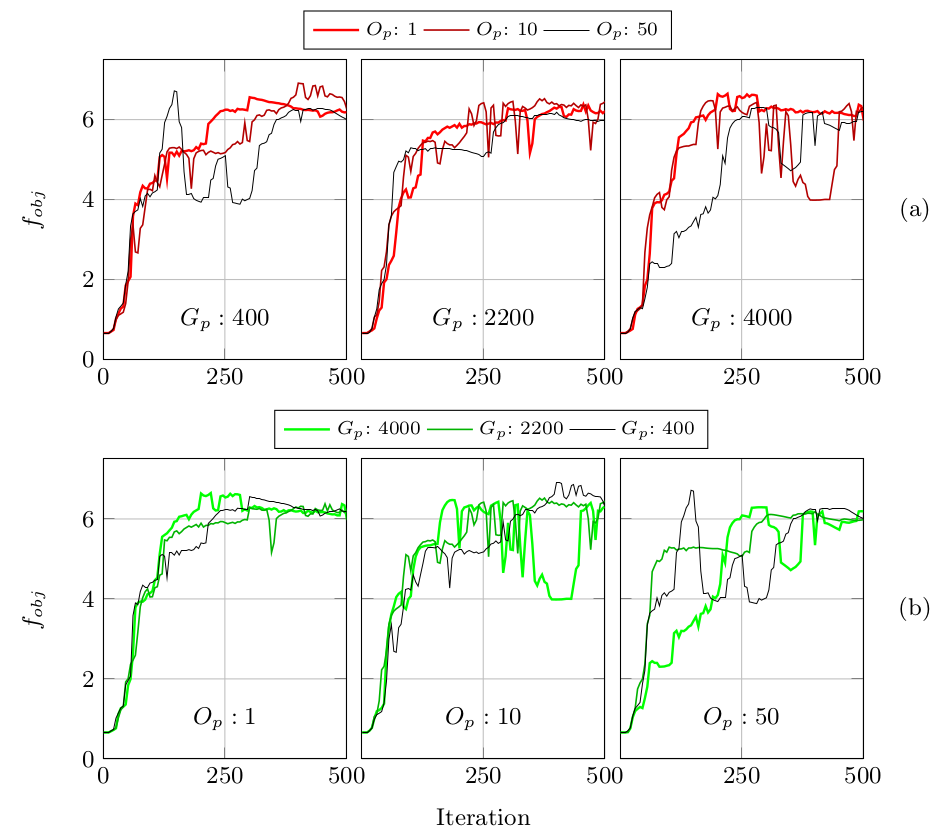}
\caption{Random search simulation objective function evolution: (a) comparison at constant generated points, (b) comparison at constant optimization period.}
\label{opt2}       
\end{figure}

According to these results, we choose \emph{random search} as the preferred optimization algorithm. Subsequently, we perform a set of simulations to investigate how the choice of the parameters influences the results. In particular, we run simulations for different numbers of generated points $G_p=\{400, 2200, 4000\}$ and for different values of the optimization period $O_p=\{1, 10, 50\}$. 
The results we obtain from these simulations are summarized in Fig.~\ref{opt2}. The sub-plots show the value of the objective function achieved in each configuration. It is possible to observe that different parameter choices provide similar results, in terms of the objective function. Therefore, we conclude that the optimization algorithm can be effectively run choosing the lowest values of generated points (i.e. $G_p=400$), and with the largest value of the optimization period (i.e. $O_p=50$): these choices reduce the computational requirements without causing a significant decrease in the quality of the achieved solution.

\section{Experimental validation}\label{sec:experiments}

\begin{figure}[!hbt]
	\centering
	\includegraphics[width=0.2\columnwidth] {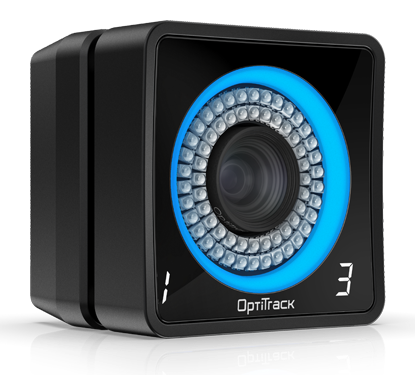}
	\hspace{2.5cm}
	\includegraphics[width=0.2\columnwidth] {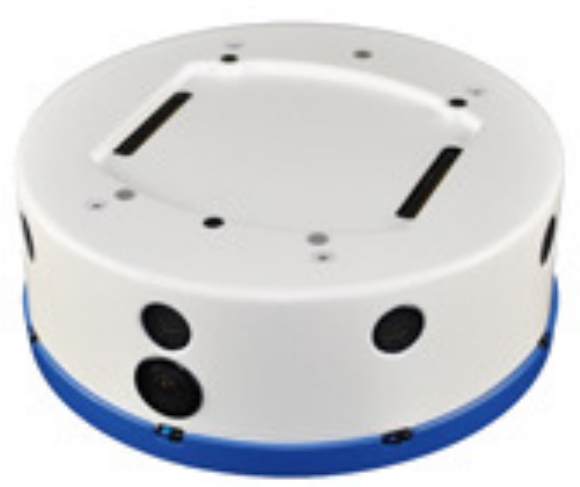}
	\caption{One of four OptiTrack Prime 13 cameras and one of eight K-Team's Khepera IV robots ({\o}$=14.0\,cm, h = 6.0\,cm$) used for the experimental setup in Section~\ref{sec:experiments}.}
	\label{img:kh4}
\end{figure}

Segueing from simulation to real robots can be challenging and results in performance degradation, especially with resource constraint hardware \cite{panerati2018}. For demonstrating the portability of the proposed online optimization, and to assess how hardware limitations affect the choice of the optimization parameters (i.e., the generated points $G_p$ and optimization period $O_p$), we transferred our methodology onto a distributed multi-robot system. The robot team consists of eight two-wheeled K-Team Khepera IV (KH4) depicted in Figure \ref{img:kh4}. Each robot comes with an 800MHz ARM Cortex-A8 and the Yocto operating system\footnote{
	\url{https://www.k-team.com/mobile-robotics-products/khepera-iv}
}.

A camera-based tracking system (OptiTrack\footnote{
	\url{https://optitrack.com/products/prime-13/specs.html}
}'s Prime13, see Figure \ref{img:kh4}), and the \texttt{blabbermouth}\footnote{
\url{https://github.com/MISTLab/blabbermouth}
} communication infrastructure are combined to emulate range and bearing sensors for each robot. This also enables point-to-point communication with a limited communication range $R$ between the robots (a similar setup was used in \cite{panerati2018}).  

The optimization procedure described in Section \ref{sec:impl_eval} is embedded into the executable \texttt{bzzkh4}\footnote{\url{https://github.com/MISTLab/BuzzKH4}} that runs the Buzz byte code of each robotic controller.
Starting from the in simulation investigated parameters, we determine the optimization times $\Delta_t$ for processing on the KH4 multi-robot system by varying $G_p$.
A set of optimizations is performed with the initial topology configuration (introduced in Section \ref{sec:impl_eval}). We obtain $\Delta_t$'s of $8'41''$, $46'47''$ and $84'23''$ as runtimes for 400, 2200 and 4000 generated points $G_p$, respectively. With increasing $G_p$, $\Delta_t$ increases linearly and ranges from  minutes to hours. Considering these computational demands, it is sensible to run the online optimization on the KH4 every $O_p = 50$ steps with $G_p = 400$ points.

\begin{figure}[!htb]
	\includegraphics[width=\columnwidth]{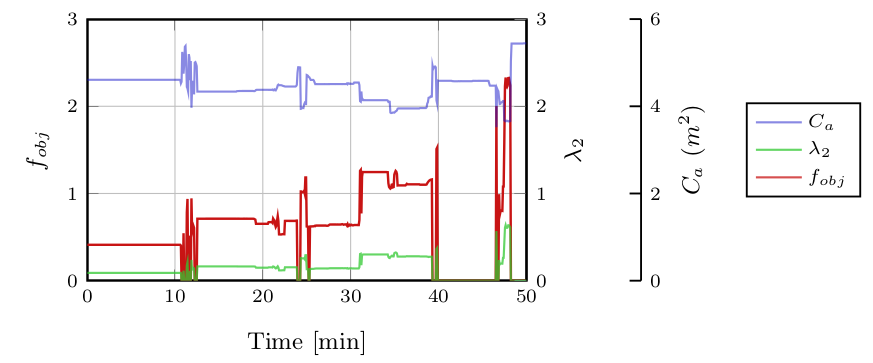}
	\caption{Objective function $f_{obj}$ evolution achieved in real robot experiments using random search optimization algorithm set with 400 $G_p$ and 50 $O_p$.
	} 
	\label{real_obj_f_2}       
\end{figure}

Simulations and experimental validation iterate over a fix number of control steps. After 500 such iterations, we consider an experiment to be finished. Due to different processing times on each member of the robot team, the KH4's operate asynchronously. Some reach the end of the experiment earlier and will thus stop communicating and end their operations. Figure \ref{real_obj_f_2} shows the evolution of three functions: objective function $f_{obj}$, the coverage $C_a$ and the connectivity $\lambda_2$. The objective function increases, which corresponds to the simulated behavior (compare to Figure \ref{opt2}). The initial decrease in coverage, as well as the increasing of connectivity have been observed in the previous experiments \cite{panerati2018}.

\section{Conclusions}\label{sec:conclusions}

{In this paper, we propose online optimization to automatically tune the gains of a control law for resilient multi-robot networks.}
{Our starting point is a control law~\cite{ghedini2017b} that was proven to increase the robustness of an initially connected multi-robot topology.}
{Here, we extend that work with the following contributions: (i) we implement an online framework to predict and optimize the multi-robot system performance; (ii) within it, we compare three different optimization algorithms; and finally, (iii) we assess the feasibility of implementing this framework on a real robotic setup comprising of eight 2-wheel robots.}
{Simulations demonstrate the effectiveness of the proposed approach as well as its low sensitivity to parameterization. }
{We also demonstrate that the methodology can be executed on robots with limited computational capabilities.}
{Future developments of this work will include the validation of our methodology using ROS-based flying robots and its study with desired control actions other than coverage.}

\bibliographystyle{abbrv}
\bibliography{biblio}



\end{document}